# PREDICTABILITY OF PV POWER GRID PERFORMANCE ON INSULAR SITES WITHOUT WEATHER STATIONS: USE OF ARTIFICIAL NEURAL NETWORKS**

C. Voyant[1,2], M. Muselli[2*], C. Paoli[2], ML Nivet[2], P Poggi[2] and P. Haurant[2]

1- Hospital of Castelluccio, radiotherapy unit, B.P.85 20177 Ajaccio - France
2- University of Corsica/CNRS UMR SPE 6134, {Rte des Sanguinaires, 20000 Ajaccio/Campus Grimaldi, 20250 Corte} - France

*: Corresponding author: phone/fax: +33(0)4 955 241 30/42, marc.muselli@univ-corse.fr
**: a part of this research is founded by CTC (Collectivité Territoriale de Corse)

ABSTRACT: The official meteorological network is poor on the island of Corsica: only three sites being about 50 km apart are equipped with pyranometers which enable measurements by hourly and daily step. These sites are Ajaccio (41°55'N and 8°48'E, seaside), Bastia (42°33'N, 9°29'E, seaside) and Corte (42°30'N, 9°15'E average altitude of 486 meters). This lack of weather station makes difficult the predictability of PV power grid performance. This work intends to study a methodology which can predict global solar irradiation using data available from another location for daily and hourly horizon. In order to achieve this prediction, we have used Artificial Neural Network which is a popular artificial intelligence technique in the forecasting domain. A simulator has been obtained using data available for the station of Ajaccio that is the only station for which we have a lot of data: 16 years from 1972 to 1987. Then we have tested the efficiency of this simulator in two places with different geographical features: Corte, a mountainous region and Bastia, a coastal region. On daily horizon, the relocation has implied fewer errors than a "naïve" prediction method based on the persistence (RMSE=1468 Vs 1383Wh/m² to Bastia and 1325 Vs 1213Wh/m² to Corte). On hourly case, the results were still satisfactory, and widely better than persistence (RMSE=138.8 Vs 109.3 Wh/m² to Bastia and 135.1 Vs 114.7 Wh/m² to Corte). The last experiment was to evaluate the accuracy of our simulator on a PV power grid localized at 10 km from the station of Ajaccio. We got errors very suitable (nRMSE=27.9%, RMSE=99.0 W.h) compared to those obtained with the persistence (nRMSE=42.2%, RMSE=149.7 W.h).

## 1 PRESENTATION AND ISSUE

We present the results of the prediction of global radiation using Artificial Neural Networks (ANN) which are a popular artificial intelligence technique in the forecasting domain [1]. Inspired by biological neural networks, researchers in a number of scientific disciplines are designing ANNs to solve a variety of problems in decision making, optimization, control and obviously prediction [2-3]. In this context, our aim was to answer to the following question: Can we design an ANN of a site for which there is a lot of solar radiation data available and use this ANN to predict a PV power grid performance of another site? We tried to answer to this question with sites located on the island of Corsica (France). The island is characterized by a Mediterranean climate and a hilly terrain. The official meteorological network (from the French Meteorological Organization) is very poor: only three sites being about 50 km apart are equipped with pyranometers and enable measurements by hourly and daily step. These sites are Ajaccio (41°55'N and 8°48'E, seaside), Bastia (42°33'N, 9°29'E, seaside) and Corte (42°30'N, 9°15'E average altitude of 486 meters). In this study, we focus on the prediction of global solar irradiation on a horizontal plane for daily and hourly horizon. These time steps have been chosen according to the electricity supplier (EDF: Electricité De France) who is interested in the estimation of the fossil fuel saving. It is very important for a remote site where electrification can be problematic [4], and for quantifying the solar potential available. Indeed, this is very important both for the power plant implementation and for sizing of PV array. Solar radiation has been measured for a long time, but even today there are many unknown characteristics of its behavior. So, it seems appropriate to develop a prediction methodology using the data available in another location in order to overcome the lack of weather station and the demand for renewable energy source on the island.

## 2 PHYSICAL PHENOMENA

There are two approaches that allow quantifying solar radiation: the "physical modeling" based on physical process occurring in the atmosphere and influencing solar radiation [5], and the "statistic solar climatology" mainly based on time series analysis [5]. We have chosen to combine these two methods to improve the quality of prediction. In this work, we have used the physical phenomena in an attempt to overcome the seasonality of the resource. When studying the solar energy on the earth's surface with time series, habitually the non-stationarity perturbs the quality of prediction. Often it's necessary to apprehend the periodic phenomenon [6,7]. In daily case, there is seasonality with annual period, and in the hourly case there is in addition daily phenomenon.

We can use the extraterrestrial global horizontal irradiance as stationarization setting. The deterministic component of the series is thus reduced, leaving more important place to the stochastic part (cloud cover). We chose a multiplicative pattern and we stationarize the time series by using the extra-terrestrial irradiation. We divide the time series by the coefficient of extraterrestrial insolation. In the hourly case, the Earth's rotation adds a new periodic component. The solar altitude angle can be easily linked to the number of photon interacting on a horizontal surface at ground level (day light). In a first approximation, it can be considered as directly proportional to the energy received. It is why we have chosen to remove the periodicity (annual and daily) dividing the time series by the coefficient of extraterrestrial insolation, but also by the sinus of the solar height. These two treatments (hourly and daily) have been used designing an ANN in order to predict the global radiation on horizontal plane.

3  SOLAR RADIATION PREDICTION FOR A SITE WITHOUT EXPERIMENTAL DATA MEASURED

ANNs are capable of capturing the characteristics of any phenomenon with a good degree of accuracy. This technology often offers an alternative to traditional physical-based models, and excels at uncovering relationships in data. It is also a powerful non-linear estimator which is recommended when the functional form between input-output is unknown or it is not well understood. The main difficulties using the ANN technology are to have enough data (number of learning elements), to be sure of the data quality and to find the best architecture of ANN. Following the literature [8], we designed and optimized a Multi-Layer Perceptron (MLP) network which is the most used of the ANN architectures. Solar radiation data from 1972 to 1987 have been used for training and testing the MLP. 80% of the data available were used to train the MLP and 20% to test the MLP. We used the Matlab software to establish our network which has 8 entries, 3 hidden neurons and 1 output. These 8 entries are the 8 previous hourly measures considered (t, t-1, t-2, .. , t-7) of global solar irradiation (Wh/m²) and the output is the measure to predict for the next hour (t+1). In daily case, hours are replaced by days. We obtained good results with a prediction and a learning done on the same site: nRMSE less than 20.3% in daily horizon, 19.5% for hourly horizon and 16 years of learning. We want to use this ANN, trained successfully on site with lots of data available (Ajaccio), to estimate the insolation on a site which has no meteorological station and therefore no measure history. Many studies have tested this kind of prediction [1,9], often by establishing a regression of parameters like latitude, longitude, altitude, days of the year, etc. Their methodology doesn't use the time series property and therefore doesn't allow predicting the stochastic cloud cover but only an insolation average. As we developed above, in addition to the data of Ajaccio, we have two meteorological stations: one in Corte (mountain location) and the other one in Bastia (coastal location). In order to give a first answer to the question of the training relocation feasibility, we proposed to test both in Bastia and Corte, three models of prediction. In the case A, we use the MLP trained with data from Ajaccio (16 years from 1972 to 1987). In the case B, we use the MLP trained with data from the selected place. We have 5 years of data in Corte (from 2002 to 2006) and also 5 years in Bastia (from 1991 to 1995). And in the case C, we use a "naïve" prediction method based on the persistence. The following tables present our results for a full year (1996 for Bastia and 2007 for Corte) of global horizontal irradiance prediction by daily (Table 1) and hourly step (Table 2).

**Table I:** Comparison of the 3 techniques of prediction for the daily forecast: horizon 1 day

| Forecast Location | Techniques of Prediction | RMSE (Wh/m²) NRMSE (%)± IC95 | CC |
|---|---|---|---|
| Bastia (year 1996) | Ajaccio (A) | 1383  29.19±0,13 | 0.842 |
| | Bastia (B) | 1288  27.51±0,2 | 0.842 |
| | Persistence (C) | 1468  31.4 | 0.807 |
| Corte (year 2007) | Ajaccio (A) | 1213  25.88±0,17 | 0.887 |
| | Corte (B) | 1112  23.73±0,15 | 0.887 |
| | Persistence (C) | 1325  28.3 | 0.844 |

On the table 1, as we could imagine, the training location has modified the quality of the prediction (error represented by Root Mean Square Error with Interval Confidence 95%, Normalised Root Mean Square Error and Correlation Coefficient). Even it's difficult to compare the Corte and Bastia prediction (the years of forecasting are not the same), we see that the relocation is partly compensated by the large period of learning available at Ajaccio (16 years). However, the relocation of learning implies fewer errors than the "naïve" predictor (RMSE=1468 Vs 1383Wh/m² to Bastia and 1325 Vs 1213Wh/m² to Corte). Also, the results of correlation coefficient do not show significant differences between cases A and B. Whatever the learning location, the Corte global radiation is more predictable than Bastia ones, which combines a particular hydrography and orography.

On Table 2, we can see the difficulty of hourly predicting on the site of Corte (nRMSE =114.7Wh/m² for relocation learning and 109,6Wh/m² for standard learning). A possible explanation could be that in mountainous region the climate change can occur very quickly, confusing an ANN learned in a coastal region. The results are still satisfactory, and widely better than those of persistence.

**Table II:** Comparison of the 3 techniques of prediction for the hourly forecast: horizon 1 hour

| Forecast Location | Techniques of Prediction | RMSE (Wh/m²) NRMSE (%)± IC95 | CC |
|---|---|---|---|
| Bastia (year 1996) | Ajaccio (A) | 109.3 22.36±0.32 | 0.916 |
| | Bastia (B) | 109.1 22.32±0.32 | 0.916 |
| | Persistence (C) | 138.8 27.22 | 0.869 |
| Corte (year 2007) | Ajaccio (A) | 114.7  23.12±0.6 | 0.916 |
| | Corte (B) | 109.6 22.08±0.32 | 0.919 |
| | Persistence (C) | 135.1 26.50 | 0.889 |

If we consider only the model of prediction A, the two tables show that the worst results are obtained in Corte (RMSE>114Wh/m²) for the hourly case, and in Bastia (RMSE>1300Wh/m²) for the daily case. The figure 1 illustrates these two cases.

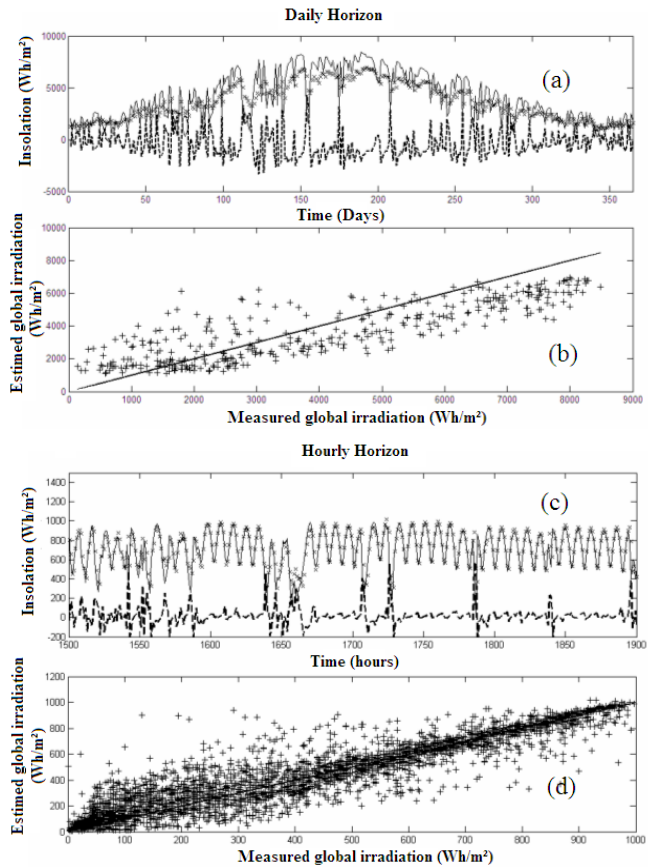

**Figure 1:** Example of simulated and experimental global solar irradiation in Bastia for the daily horizon (a) and in Corte for the hourly horizon (c). Dashed line is the error, solid line is the measured insolation and marks are the prediction. On the bottom corresponds to the correlation between experimental and simulated global irradiation in Bastia for the daily horizon (b) and in Corte for the hourly horizon (d).

On part (a) and (b), forecasting is done during the year 1996, and on part (c) and (d) on the year 2007. We can see that the ANN cannot predict atypical insolation (picks of the error dashed line) as the day with a lots of cloud cover. These phenomena are compounded by the relocation, but they also exist when learning and prediction are made on the same site. On the Figure 1.b, we can see that the relocation proposed with the Bastia's station suffers of an adjustment problem. Indeed, it overestimates the weak measures, and we underestimate the highest. This phenomenon is probably due to the normalization of data during the simulation with ANN. This normalization does not include any previous data from Bastia (assumed unknown), and is made in relation to the maximum and minimum insolation of Ajaccio. This extremum can be different on the two locations. The hourly horizon shows a great similarity between the measurement and prediction (Figure 1.c-d), this kind of prediction is relatively well supported by delocalized learning.

## 4 APPLICATION OF THE RELOCATION METHOD FOR A PV SYSTEM

A frontage PV system has been installed recently in our laboratory (Ajaccio). It has a nominal power of 6.525 kW composed by respectively 1.8 kW and 4.725 kW amorphous and mono-crystal PV modules built in 6 independent power subsystems. PV power predictions from ANN methodology described in this paper have been computed from one of this whole PV plant on a frontage side exposed to the south (azimuth null) and tilted at 80°. The PV system is composed by 9 SUNTECH 175S-24Ac for a 1.175 kW nominal power connected to a 1.85 kW SUNNY BOY SMA inverter for PV production on the grid. For the PV power calculation, we use in first approximation, a linear production based on a constant PV plant efficiency $\eta_{PV}$ = 13%, measured from a PV-KLA I-V curve plotter device (INGENIEURBÜRO Mencke & Tegtmeyer):

$E_{PV}$ (Wh) = $\eta_{PV}$ $I_\beta$ S.

$I_\beta$ is the hourly global irradiation on the PV system ($\beta$ = 80°) estimated from ANN, and S is the usable surface of the PV system under consideration (S = 10.125 m²). To predict this energy, we have used the 16 years of global horizontal irradiation available on the site of Ajaccio, like learning set. Classical models are used to compute tilted irradiation for an 80° angle:

$I_\beta = I_{\beta=0} [ I_\beta / I_{\beta=0} ]^{clear\ sky}$.

The last term of this equation was calculated from the software PVSYST [10]. The ANN designing previously has been used this time to predict the energy produced by the PV plant of our laboratory. The distance is about 10 km from the place of obtaining the series of training.

The results presented in Figure 2 seem very promising, but can not completely validate the methodology, because of the limited number of comparable events. It is interesting to note that in the case with a simulated ANN and learning delocalized, we get errors very suitable (nRMSE=27.9%, RMSE=99 Wh) compared to those obtained with the persistence (nRMSE=42.2%, RMSE=149.7 Wh).

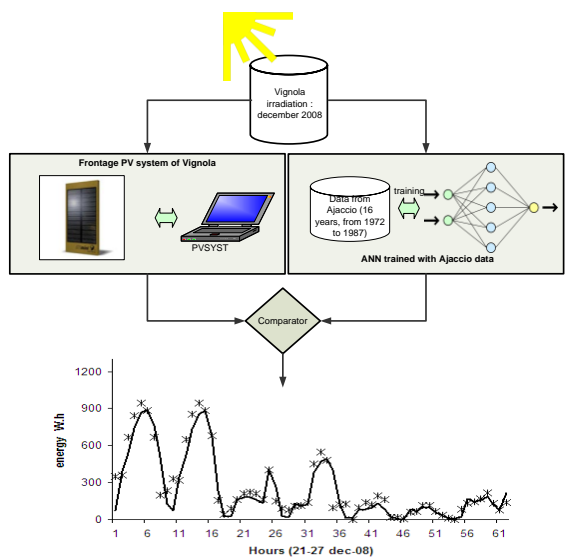

**Figure 2:** Comparison between computed and measured energy generated with the frontage PV system used in experimentation.

## 5 CONCLUSION

In this work we have studied a methodology which can predict global solar irradiation using data available from another location for daily and hourly horizon. In order to achieve this prediction, we have used an ANN. In the case of global horizontal radiation, the outsourced learning solution is possible in the hourly case: errors are almost identical to the case of a localized learning with a five years history. Our forecast methodology allows studying many sites with no meteorological station. The results show that if the hourly horizon is well suited to this relocation, the daily horizon has yet to be studied, to consider a generalization of the relocation and the versatility of ANN. The prediction on Corte (mountain town), where environmental constraints are not at all that we can have in Ajaccio, is very reliable. For Bastia (a seaside town like Ajaccio), the results are very promising but requires further investigation (especially in the daily case). The results found in this study are very interesting because they show "hidden links" between the three sites. The results of the prediction of PV systems, shown that ANN and delocalized learning may be very useful, but need to increase and finalize the comparison between measurement and simulation. In next works, we try to improve our model, and validate our approach with solar module installed on our laboratory (Vignola, Ajaccio). At term, this tool (ANN) could eventually help the system manager, for implanting new PV central and so not limit the abundance to the sites with weather stations.